\documentclass{llncs}
\input glyphtounicode\pdfgentounicode=1

\pdfoutput=1

\usepackage{amsmath}                
\usepackage{float}                  
\usepackage{graphicx}               
\usepackage{epstopdf}               
\usepackage[perpage]{footmisc}      
\usepackage{nicefrac}               
\usepackage{multirow,bigdelim}       
\usepackage{booktabs}               
\usepackage{nicefrac}
\usepackage{bm}
\usepackage{multirow}
\usepackage{colortbl}
\usepackage{subcaption}
\begin{document}
\raggedbottom

\title{Uncertainty in multitask learning: joint representations for probabilistic MR-only radiotherapy planning}
\titlerunning{Probabilistic multitask learning}
\author{Felix J.S. Bragman\inst{1}, Ryutaro Tanno\inst{1}, Zach Eaton-Rosen\inst{1}, \\ Wenqi Li\inst{1}, David J. Hawkes\inst{1}, Sebastien Ourselin\inst{2} \\ Daniel C. Alexander\inst{1,3}, Jamie R. McClelland\inst{1} \and M. Jorge Cardoso\inst{2,1}}
\index{Bragman, Felix}
\index{Tanno, Ryutaro}
\index{Eaton-Rosen, Zach}
\index{Li, Wenqi}
\index{Hawkes, David}
\index{Ourselin, Sebastien}
\index{Alexander, Daniel}
\index{McClelland, Jamie}
\index{Cardoso, M. Jorge}
\institute{Centre for Medical Image Computing, University College London, UK
\and Biomedical Engineering and Imaging Sciences, King's College London, UK
\and Clinical Imaging Research Centre, National University of Singapore, Singapore}
\maketitle

\begin{abstract}
Multi-task neural network architectures provide a mechanism that jointly integrates information from distinct sources. It is ideal in the context of MR-only radiotherapy planning as it can jointly regress a synthetic CT (synCT) scan and segment organs-at-risk (OAR) from MRI. We propose a probabilistic multi-task network that estimates: 1) \textit{intrinsic} uncertainty through a heteroscedastic noise model for spatially-adaptive task loss weighting and 2) \textit{parameter} uncertainty through approximate Bayesian inference. This allows sampling of multiple segmentations and synCTs that share their network representation. We test our model on prostate cancer scans and show that it produces more accurate and consistent synCTs with a better estimation in the variance of the errors, state of the art results in OAR segmentation and a methodology for quality assurance in radiotherapy treatment planning.
\end{abstract}

\section{Introduction}
Radiotherapy treatment planning (RTP) requires a magnetic resonance (MR) scan to segment the target and organs-at-risk (OARs) with a registered computed tomography (CT) scan to inform the photon attenuation. MR-only RTP has recently been proposed to remove dependence on CT scans as cross-modality registration is error prone whilst extensive data acquisition is labourious. MR-only RTP involves the generation of a synthetic CT (synCT) scan from MRI. This synthesis process, when combined with manual regions of interest and safety margins provides a deterministic plan that is dependent on the quality of the inputs. Probabilistic planning systems conversely allow the implicit estimation of dose delivery uncertainty through a Monte Carlo sampling scheme. A system that can sample synCT and OAR segmentations would enable the development of a fully end-to-end uncertainty-aware probabilistic planning system. 
 
Past methods for synCT generation and OAR segmentation stem from multi-atlas propagation \cite{ninon2017}. Applications of convolutional neural networks (CNNs) to CT synthesis from MRI have recently become a topic of interest \cite{nie2017,wolterink2017}. Conditional generative adversarial networks have been used to capture fine texture details \cite{nie2017} whilst a CycleGAN has been exploited to leverage the abundance of unpaired training sets of CT and MR scans \cite{wolterink2017}. These methods however are fully deterministic. In a probabilistic setting, knowledge of the posterior over the network weights would enable sampling multiple realisations of the model for probabilistic planning whilst uncertainty in the predictions would be beneficial for quality control. Lastly, none of the above CNN methods segment OARs. If a model were trained in a multi-task setting, it would produce OAR segmentations and a synCT that are anatomically consistent, which is necessary for RTP. \looseness=-1  

Past approaches to multi-task learning have relied on uniform or hand-tuned weighting of task losses \cite{moeskops2016deep}. Recently, Kendall et al. \cite{kendall2017multi} interpreted homoscedastic uncertainty as task-dependent weighting. However, homoscedastic uncertainty is constant in the task output and unrealistic for imaging data whilst yielding non-meaningful measures of uncertainty. Tanno et al. \cite{tanno2017bayesian} and Kendall et al. \cite{kendall2017uncertainties} have raised the importance of modelling both \textit{intrinsic} and \textit{parameter} uncertainty to build more robust models for medical image analysis and computer vision. \textit{Intrinsic} uncertainty captures uncertainty inherent in observations and can be interpreted as the irreducible variance that exists in the mapping of MR to CT intensities or in the segmentation process. \textit{Parameter} uncertainty quantifies the degree of ambiguity in the model parameters given the observed data. 

This paper makes use of \cite{tanno2017bayesian} to enrich the multi-task method proposed in \cite{kendall2017multi}. This enables modelling the spatial variation of \textit{intrinsic} uncertainty via heteroscedastic noise across tasks and integrating \textit{parameter} uncertainty via dropout \cite{gal2016dropout}. We propose a probabilistic dual-task network, which operates on an MR image and simultaneously provides three valuable outputs necessary for probabilistic RTP: (1) synCT generation, (2) OAR segmentation and (3) quantification of predictive uncertainty in (1) and (2) (Fig.\ref{fig:diagram1}). The architecture integrates the methods of uncertainty modelling in CNNs \cite{tanno2017bayesian,kendall2017uncertainties} into a multi-task learning framework with hard-parameter sharing, in which the initial layers of the network are shared across tasks and branch out into task-specific layers (Fig.\ref{fig:diagram}). Our probabilistic formulation not only provides an estimate of uncertainty over predictions from which one can stochastically sample the space of solutions, but also naturally confers a mechanism to spatially adapt the relative weighting of task losses on a voxel-wise basis.

\section{Methods}
We propose a probabilistic dual-task CNN algorithm which takes an MRI image, and simultaneously estimates the distribution over the corresponding CT image and the segmentation probability of the OARs. We use a heteroscedastic noise model and binary dropout to account for \textit{intrinsic} and \textit{parameter} uncertainty, respectively, and show that we obtain not only a measure of uncertainty over prediction, but also a mechanism for data-driven spatially adaptive weighting of task losses, which is integral in a multi-task setting. We employ a patch-based approach to perform both tasks, in which the input MR image is split into smaller overlapping patches that are processed independently. For each input patch $\mathbf{x}$, our dual-task model estimates the conditional distributions $p(\mathbf{y}_i|\mathbf{x})$ for tasks $i=1,2$ where $\mathbf{y}_1$ and $\mathbf{y}_2$ are the Hounsfield Unit and OAR class probabilities. At inference, the probability maps over the synCT and OARs are obtained by stitching together outputs from appropriately shifted versions of the input patches.\looseness=-1

\subsubsection{Dual-task architecture.} 
\begin{figure}[!t]
        \vspace{-5mm}
		\centering
\includegraphics[height=0.185\textwidth]{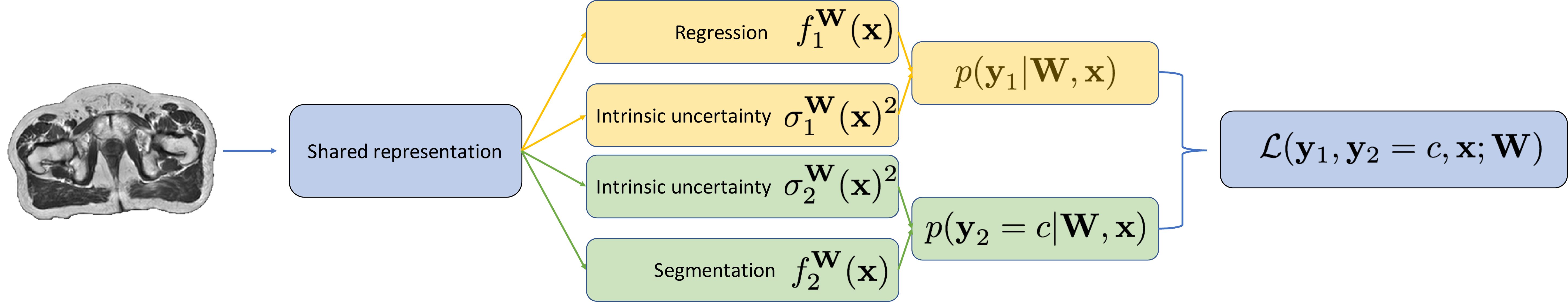}
		\caption{Multi-task learning architecture. The predictive mean and variance $[f^{\mathbf{W}}_i(\text{\textbf{x}}), \sigma_{i}^{\mathbf{W}}(\text{\textbf{x}})^{2}]$ are estimated for the regression and segmentation. The task-specific likelihoods p$\left(\text{\textbf{y}}_{i}|\text{\textbf{W}},\text{\textbf{x}}\right)$ are combined to yield the multi-task likelihood p$\left(\text{\textbf{y}}_{1}, \text{\textbf{y}}_{2}|\text{\textbf{W}},\mathbf{x}\right)$.} 
		\label{fig:diagram}
        \vspace{-5mm}
	\end{figure}

We perform multi-task learning with hard-parameter sharing \cite{Caruana1993MultitaskLA}. The model shares the initial layers across the two tasks to learn an invariant feature space of the anatomy and branches out into four task-specific networks with separate parameters (Fig.\ref{fig:diagram}). There are two networks for each task (regression and segmentation). where one aims to performs CT synthesis (regression) or OAR segmentation, and the remaining models \emph{intrinsic} uncertainty associated to the data and the task.\looseness=-1
    
The rationale behind shared layers is to learn a joint representation between two tasks to regularise the learning of features for one task by using cues from the other. We used a high-resolution network architecture (HighResNet) \cite{li2017compactness} as the shared trunk of the model for its compactness and accuracy shown in brain parcellation. HighResNet is a fully convolutional architecture that utilises dilated convolutions with increasing dilation factors and residual connections to produce an end-to-end mapping from an input patch (\textbf{x}) to voxel-wise predictions (\textbf{y}). The final layer of the shared representation is split into two task-specific compartments (Fig. \ref{fig:diagram}). Each compartment consists of two fully convolutional networks which operate on the output of representation network and together learn task-specific representation and define likelihood function $p\left(\text{\textbf{y}}_{i}|\text{\textbf{W}},\textbf{x}\right)$ for each task $i=1, 2$ where \text{\textbf{W}} denotes the set of all parameters of the model. 

\subsubsection{Task weighting with heteroscedastic uncertainty.}
Previous probabilistic multitask methods in deep learning \cite{kendall2017multi} assumed constant intrinsic uncertainty per task. In our context, this means that the inherent ambiguity present across synthesis or segmentation does not depend on the spatial locations within an image. This is a highly unrealistic assumption as these tasks can be more challenging on some anatomical structures (e.g. tissue boundaries) than others. To capture potential spatial variation in intrinsic uncertainty, we adapt the \emph{heteroscedastic} (data-dependent) noise model to our multitask learning problem.

For the CT synthesis task, we define our likelihood as a normal distribution $p\left(\text{\textbf{y}}_{1}|\mathbf{W},\mathbf{x}\right) = \mathcal{N}(f^{\text{\textbf{W}}}_1(\text{\textbf{x}}), \sigma_{1}^{\text{\textbf{W}}}(\text{\textbf{x}})^{2})$
where mean $f^{\text{\textbf{W}}}_1(\text{\textbf{x}})$ and variance $\sigma_{1}^{\text{\textbf{W}}}(\text{\textbf{x}})^{2}$ are modelled by the regression output and uncertainty branch as functions of the input patch $\mathbf{x}$ (Fig.\ref{fig:diagram}). We define the task loss for CT synthesis to be the negative log-likelihood (NLL) $\mathcal{L}_{1}(\text{\textbf{y}}_{1},\mathbf{x};\mathbf{W})= \frac{1}{2\sigma_{1}^{\text{\textbf{W}}}(\text{\textbf{x}})^{2}}||\text{\textbf{y}}_{1}-f^{\text{\textbf{W}}}_1(\text{\textbf{x}})||^{2} + \text{log}\sigma_{1}^{\text{\textbf{W}}}(\text{\textbf{x}})^{2}$. This loss encourages assigning high-uncertainty to regions of high errors, enhancing the robustness of the network against noisy labels and outliers, which are prevalent at organ boundaries especially close to the bone.

For the segmentation, we define the classification likelihood as softmax function of scaled logits i.e. $p\left(\text{\textbf{y}}_{2}|\textbf{W},\mathbf{x}\right) = \text{Softmax}\big{(}f_{2}^{\text{\textbf{W}}}(\text{\textbf{x}})/2\sigma^{\mathbf{W}}_{2}(\text{\textbf{x}})^2\big{)}$ where the segmentation output $f_{2}^{\text{\textbf{W}}}(\text{\textbf{x}})$ is scaled by the uncertainty term $\sigma^{\mathbf{W}}_{2}(\text{\textbf{x}})^2$ before softmax (Fig.\ref{fig:diagram}). As the uncertainty $\sigma^{\mathbf{W}}_{2}(\text{\textbf{x}})^{2}$ increases, the Softmax output approaches a uniform distribution, which corresponds to the maximum entropy discrete distribution. We simplify the scaled Softmax likelihood by considering an approximation in \cite{kendall2017multi}, $\frac{1}{\sigma^{2}}\sum_{c'}\text{exp}(\frac{1}{2\sigma^{\mathbf{W}}_{2}(\text{\textbf{x}})^2}f^{\text{\textbf{W}}}_{2,c'}(\text{\textbf{x}})) \approx \left(\sum_{c'}\text{exp}(f^{\mathbf{W}}_{2,c'}(\text{\textbf{x}}))\right)^{1/2\sigma^{\mathbf{W}}_{2}(\text{\textbf{x}})^2}$ where $c'$ denotes a segmentation class. This yields the NLL task-loss of the form $\mathcal{L}_{2}(\text{\textbf{y}}_{2}=c,\mathbf{x};\mathbf{W}) \approx \frac{1}{2\sigma^{\mathbf{W}}_{2}(\text{\textbf{x}})^2} \text{CE}(f_{2}^{\text{\textbf{W}}}(\text{\textbf{x}}),\text{\textbf{y}}_{2}=c)+ \text{log}\sigma^{\mathbf{W}}_{2}(\text{\textbf{x}})^2$, where CE denotes cross-entropy.

The joint likelihood factorises over tasks such that $p\left(\text{\textbf{y}}_{1},\text{\textbf{y}}_{2}|\mathbf{W},\mathbf{x}\right)=\prod^{2}_{i}p\left(\text{\textbf{y}}_{i}|\mathbf{W},\mathbf{x}\right)$. We can therefore derive the NLL loss for the dual-task model as
\begin{equation*}
\mathcal{L}(\textbf{y}_{1},\textbf{y}_{2}=c,\mathbf{x};\mathbf{W}) =\frac{||\text{\textbf{y}}_{1}-f^{\text{\textbf{W}}}_1(\text{\textbf{x}})||^{2}}{2\sigma_{1}^{\text{\textbf{W}}}(\text{\textbf{x}})^{2}} + \frac{\text{CE}(f_{2}^{\text{\textbf{W}}}(\text{\textbf{x}}),\text{\textbf{y}}_{2}=c)}{2\sigma^{\mathbf{W}}_{2}(\text{\textbf{x}})^2} + \text{log}\Big{(}\sigma^{\mathbf{W}}_{1}(\text{\textbf{x}})^2\sigma^{\mathbf{W}}_{2}(\text{\textbf{x}})^2\Big{)}\;
\end{equation*}

where both task losses are weighted by the inverse of heteroscedastic intrinsic uncertainty terms $\sigma^{\mathbf{W}}_{i}(\text{\textbf{x}})^2$, that enables automatic weighting of task losses on a per-sample basis. The log-term controls the spread. 

\subsubsection{Parameter uncertainty with approximate Bayesian inference.} In data-scarce situations, the choice of best parameters is ambiguous, and resorting to a single estimate without regularisation often leads to overfitting. Gal et al.\cite{gal2016dropout} have shown that dropout improves the generalisation of a NN by accounting for \textit{parameter} uncertainty through an approximation of the posterior distribution over its weights $q(\mathbf{W}) \approx (\mathbf{W}|\mathbf{X},\mathbf{Y_1},\mathbf{Y_2})$ where $\mathbf{X} = \{\mathbf{x}^{(1)}, ..., \mathbf{x}^{(N)}\}$, $\mathbf{Y}_1 = \{\mathbf{y}_1^{(1)}, ..., \mathbf{y}_1^{(N)}\}$, $\mathbf{Y}_2	 = \{\mathbf{y}_2^{(1)}, ..., \mathbf{y}_2^{(N)}\}$ denote the training data. We also use binary dropout in our model to assess the benefit of modelling parameter uncertainty in the context of our multitask learning problem.

During training, for each input (or minibatch), network weights are drawn from the approximate posterior $w' \sim q(\textbf{\text{W}})$ to obtain the multi-task output $\textbf{f}^{w'}(\text{\textbf{x}}):=[f^{w'}_1(\text{\textbf{x}}),f^{w'}_2(\text{\textbf{x}}), \sigma_{1}^{w'}(\text{\textbf{x}})^{2},\sigma_{2}^{w'}(\text{\textbf{x}})^{2}]$. At test time, for each input patch $\textbf{\text{x}}$ in an MR scan, we collect output samples $\{\textbf{f}^{\,w^{(t)}}(\text{\textbf{x}})\}_{t=1}^{\text{T}}$ by performing $T$ stochastic forward-passes with $\{w^{(t)}\}_{t=1}^{\text{T}}\sim q(\mathbf{W})$. For the regression, we calculate the expectation over the $T$ samples in addition to the variance, which is the \emph{parameter} uncertainty. For the segmentation, we compute the expectation of class probabilities to obtain the final labels whilst \emph{parameter} uncertainty in the segmentation is obtained by considering variance of the stochastic class probabilities on a class basis. The final predictive uncertainty is the sum of the \emph{intrinsic} and \emph{parameter} uncertainties.

\subsubsection{Implementation details}\label{sec:imp_details} 
We implemented our model within the NiftyNet framework \cite{niftynet17} in TensorFlow. We trained our model on randomly selected $152 \times 152$ patches from 2D axial slices and reconstructed the 3D volume at test time. The representation network was composed of a convolutional layer followed by $3$ sets of twice repeated dilated convolutions with dilation factors $[1,2,4]$ and a final convolutional layer. Each layer ($l$) used a $3 \times 3$ kernel with features $f_{R}=[64, 64, 128, 256, 2048]$. Each task-specific branch was a set of $5$ convolutional layers of size $[256_{l=1, 2, 3, 4},n_{i,l=5}]$ where $n_{i,l=5}$ is equal to $1$ for regression and $\bm{\sigma}$ and equal to the number of segmentation classes. The first two layers were $3 \times 3$ kernels whilst the final convolutional layers were fully connected. A Bernouilli drop-out mask with probability $p=0.5$ was applied on the final layer of the representation network. We minimised the loss using ADAM with a learning rate $10^{-3}$ and trained up to $19000$ iterations with convergence of the loss starting at 17500. For the stochastic sampling, we performed model inference $10$ times at iterations $18000$ and $19000$ leading to a set of $T=20$ samples. \looseness=-1

\section{Experiments and Results}

\subsubsection{Data}
We validated on $15$ prostate cancer patients, who each had a T2-weighted MR (3T, $1.46\times1.46\times5$mm$^{3}$) and CT scan (140kVp, $0.98\times0.98\times1.5$mm$^{3}$) acquired on the same day. Organ delineation was performed by a clinician with labels for the left and right femur head, bone, prostate, rectum and bladder. Images were resampled to isotropic resolution. The CT scans were spatially aligned with the T2 scans prior to training \cite{ninon2017}. In the segmentation, we predicted labels for the background, left/right femur head, prostate, rectum and bladder.

\subsubsection{Experiments}
We performed 3-fold cross-validation and report statistics over all hold-out sets. We considered the following models: 1) baseline networks for regression/segmentation (M1), 2) baseline network with drop-out (M2a), 3) the baseline with drop-out and heteroscedastic noise (M2b), 4) multi-task network using homoscedastic task weighting (M3) \cite{kendall2017multi} and 5) multi-task network using task-specific heteroscedastic noise and drop-out (M4). The baseline networks used only the representation network with $\nicefrac{1}{2}f_{R}$ and a fully-connected layer for the final output to allow a fair comparison between single and multi-task networks. We also compared our results against the current state of the art in atlas propagation (AP) \cite{ninon2017}, which was validated on the same dataset.\looseness=-1 

\begin{figure}[!t]
	\centering
	{\includegraphics[height=0.21\textwidth]{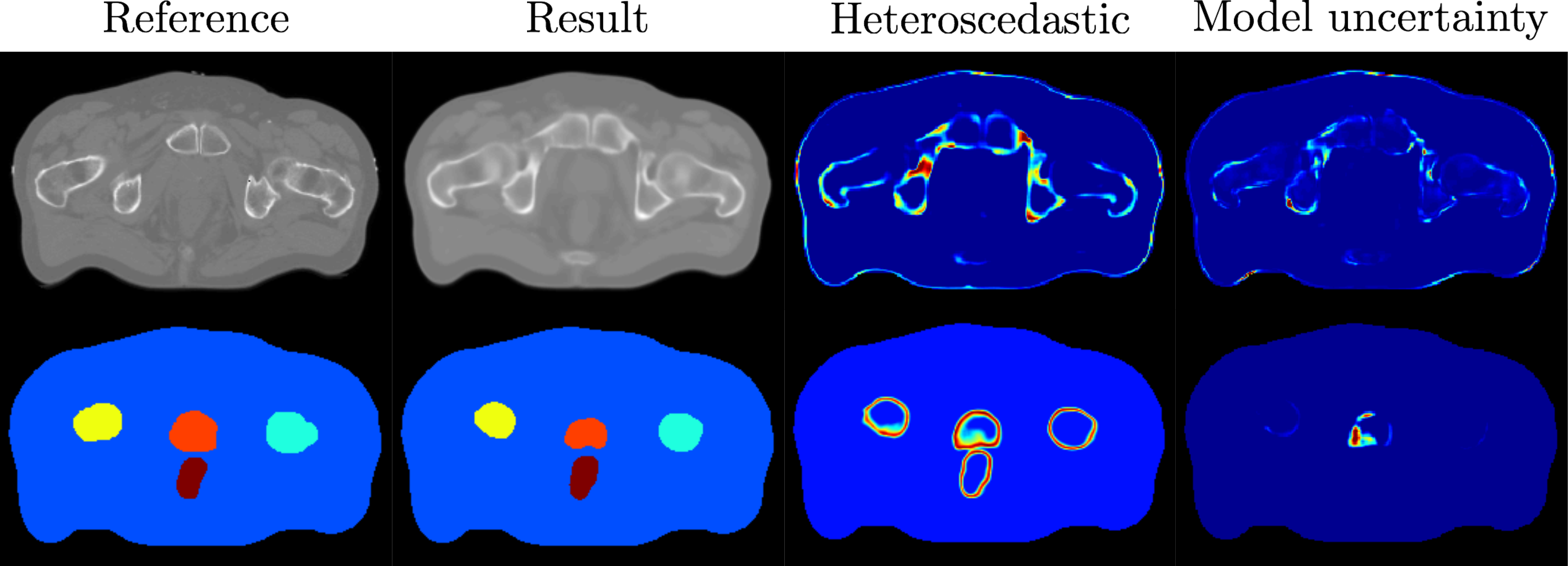}}%
	\hspace{3pt}
	{\includegraphics[height=0.21\textwidth]{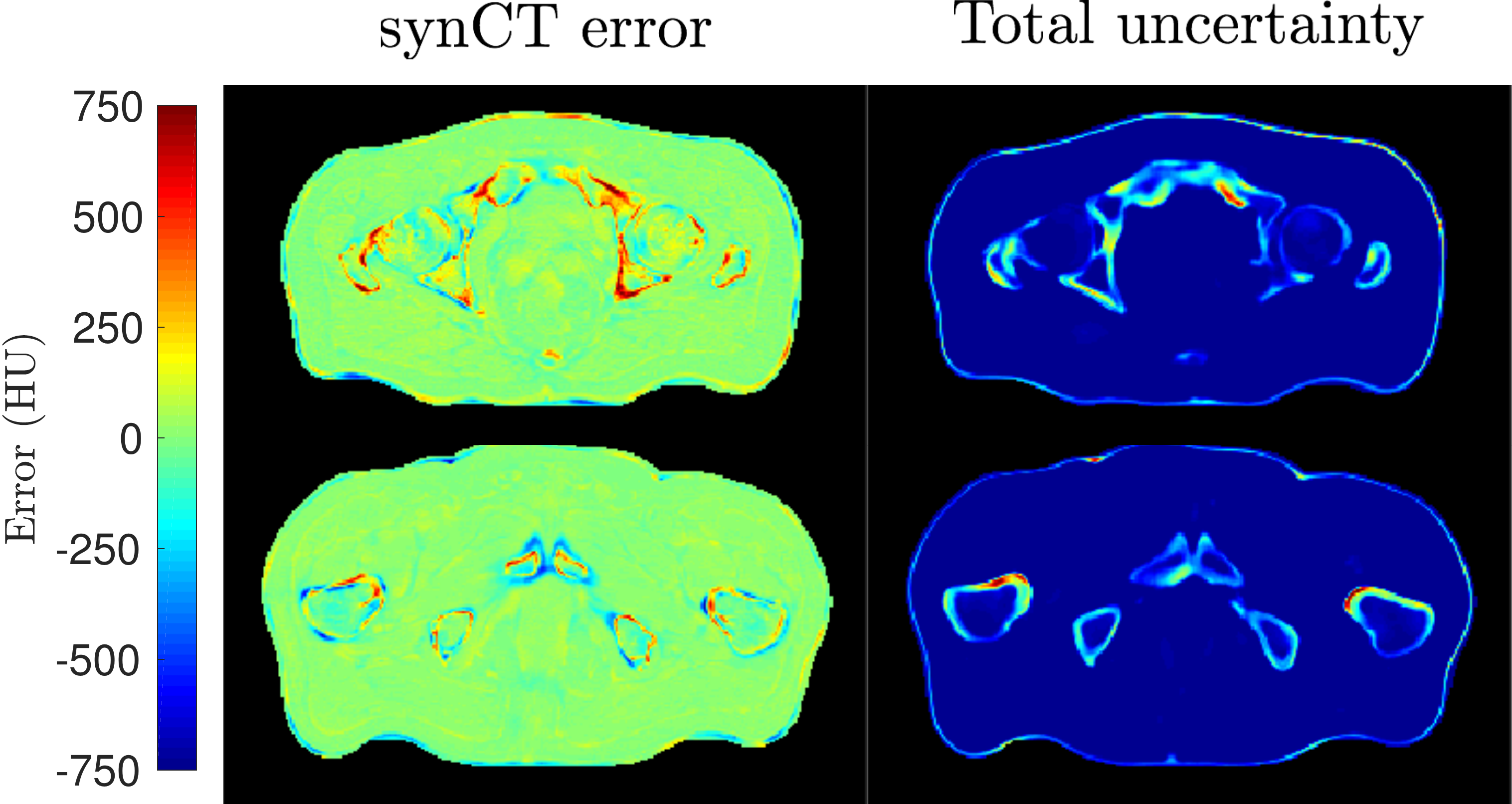}}%
       \vspace{-1mm}
\caption{Model output. \textit{Intrinsic} and \textit{parameter} uncertainty both correlate with regions of high contrast (bone in the regression, organ boundary for segmentation). Note the correlation between model error and the predicted uncertainty.}
\label{fig:diagram1}
\vspace{-5mm}
\end{figure}

\subsubsection{Model performance}
\begin{table}[!b]
\caption{Model comparison. Bold values indicate when a model was significantly worse than M4 $p<0.05$. No data was available for significance testing with AP. M2b was statistically better $p<0.05$ than M4 in the prostate segmentation.}
\begin{center}
\scalebox{0.80}{
\begin{tabular}{lccccccc}
\toprule
        Models & All & Bone & $L$ femur & $R$ femur & Prostate & Rectum & Bladder \\
\midrule
\multicolumn{8}{c}{Regression - synCT - Mean Absolute Error (HU)} \\
\midrule
M1 			   & \textbf{48.1(4.2)} & \textbf{131(14.0)}  & 78.6(19.2) & \textbf{80.1(19.6)}   & \textbf{37.1(10.4)}  & 63.3(47.3)  & \textbf{24.3(5.2)} \\
M2a 			   & \textbf{47.4(3.0)} & \textbf{130(12.1)}  & 78.0(14.8) & 77.0(13.0)   & \textbf{36.5(7.8)}  & 67(44.6)  & \textbf{24.1(7.5)} \\
M2b \cite{kendall2017uncertainties} & 44.5(3.6) & 128(17.1)  & 75.8(20.1) & 74.2(17.4)   & 31.2(7.0)  & 56.1(45.5)  & 17.8(4.7) \\
M3 \cite{kendall2017multi}& 44.3(3.1) & 126(14.4)  & 74.0(19.5) & 73.7(17.1)   & 29.4(4.7)  & 58.4(48.0) & 18.2(3.5) \\
AP \cite{ninon2017}& 45.7(4.6) & 125(10.3) & - & - & - & - & - \\
M4 (ours) 			   & 43.3(2.9) & 121(12.6)  & 69.7(13.7) & 67.8(13.2)   &28.9(2.9)  & 55.1(48.1) & 18.3(6.1) \\
\midrule
\multicolumn{8}{c}{Segmentation - OAR - Fuzzy DICE score} \\
\midrule
M1 & - & - & 0.91(0.02) & 0.90(0.04) & 0.67(0.12) & 0.70(0.15) & 0.92(0.05) \\
M2a & - & - & 0.85(0.03) & 0.90(0.04) & 0.66(0.12) & 0.69(0.13) & 0.90(0.07) \\
M2b \cite{kendall2017uncertainties} & - & - & 0.92(0.02) & 0.92(0.01) & 0.77(0.07) & 0.74(0.13) & 0.92(0.03) \\
M3 \cite{kendall2017multi}& - & - & 0.92(0.02) & 0.92(0.02) & 0.73(0.07) & 0.76(0.10) & 0.93(0.02) \\
AP \cite{ninon2017} & - & - & 0.89(0.02) & 0.90(0.01) & 0.73(0.06) & 0.77(0.06) & 0.90(0.03) \\
M4 (ours) & - & - & 0.91(0.02) & 0.91(0.02) & 0.70(0.06) & 0.74(0.12) & 0.93(0.04) \\
\bottomrule
\end{tabular}}
\end{center}
\label{tab:table}
\end{table}

An example of the model output is shown in Fig \ref{fig:diagram1}. We calculated the Mean Absolute Error (MAE) between the predicted and reference scans across the body and at each organ (Tab. \ref{tab:table}). The fuzzy DICE score between the probabilistic segmentation and the reference was calculated for the segmentation (Tab. \ref{tab:table}). Best performance was in our presented method (M4) for the regression across all masks except at the bladder. Application of the multi-task heteroscedastic network with drop-out (M4) produced the most consistent synCT across all models with the lowest average MAE and the lowest variation across patients ($43.3\pm2.9$ versus $45.7\pm4.6$ \cite{ninon2017} and $44.3\pm3.1$ \cite{kendall2017multi}). This was significant lower when compared to M1 ($p<0.001$) and M2 ($p<0.001$). This was also observed at the bone, prostate and bladder ($p<0.001$). Whilst differences at $p<0.05$ were not observed versus M2b and M3, the consistent lower MAE and standard deviation across patients in M4 demonstrates the added benefit of modelling heteroscedastic noise and the inductive transfer from the segmentation task. We performed better than the current state of the art in atlas propagation, which used both T1 and T2-weighted scans \cite{ninon2017}. Despite equivalence with the state of the art (Tab. \ref{tab:table}), we did not observe any significant differences between our model and the baselines despite an improvement in mean DICE at the prostate and rectum ($0.70\pm0.06$ and $0.74\pm0.12$) versus the baseline M1 ($0.67\pm0.12$, $0.70\pm0.15$). The \emph{intrinsic uncertainty} (Fig. \ref{fig:diagram1}) models the uncertainty specific to the data and thus penalises regions of high error leading to an under-segmentation yet with higher confidence in the result.\looseness=-1  

\subsubsection{Uncertainty estimation for radiotherapy}
We tested the ability of the proposed network to better predict associated uncertainties in the synCT error. To verify that we produce clinically viable samples for treatment planning, we quantified the distribution of regression z-scores for the multi-task heteroscedastic and homoscedastic models. In the former, the total predictive uncertainty is the sum of the \emph{intrinsic} and \emph{parameter} uncertainties. This leads to a better approximation of the variance in the model. In contrast, the total uncertainty in the latter reduces to the variance of the stochastic test-time samples. This is likely to lead to a miscalibrated variance. A $\chi^{2}$ goodness of fit test was performed, showing that the homoscedastic z-score distribution is not normally distributed ($0.82 \pm 0.54$, $p<0.01$) in contrast to the heteroscedastic model ($0.04 \pm 0.84$, $p>0.05$). This is apparent in Fig.3 where there is greater confidence in the synCT produced by our model in contrast the homoscedastic case.\looseness=-1

The predictive uncertainty can be exploited for quality assurance (Fig. \ref{fig:diagram4}). There may be issues whereupon time differences have caused variations in bladder and rectum filling across MR and CT scans causing patient variability in the training data. This is exemplified by large errors in the synCT at the rectum (Fig. \ref{fig:diagram4}) and quantified by large localised z-scores (Fig. \ref{fig:diagram4}g), which correlate strongly with the \emph{intrinsic} and \emph{parameter} uncertainty across tasks (Fig. \ref{fig:diagram1} and \ref{fig:diagram4}).\looseness=-1

\begin{figure}[!t]
  \centering
  {\includegraphics[trim=0mm 0mm 0mm 0mm,clip=true,height=0.29\textwidth,keepaspectratio]{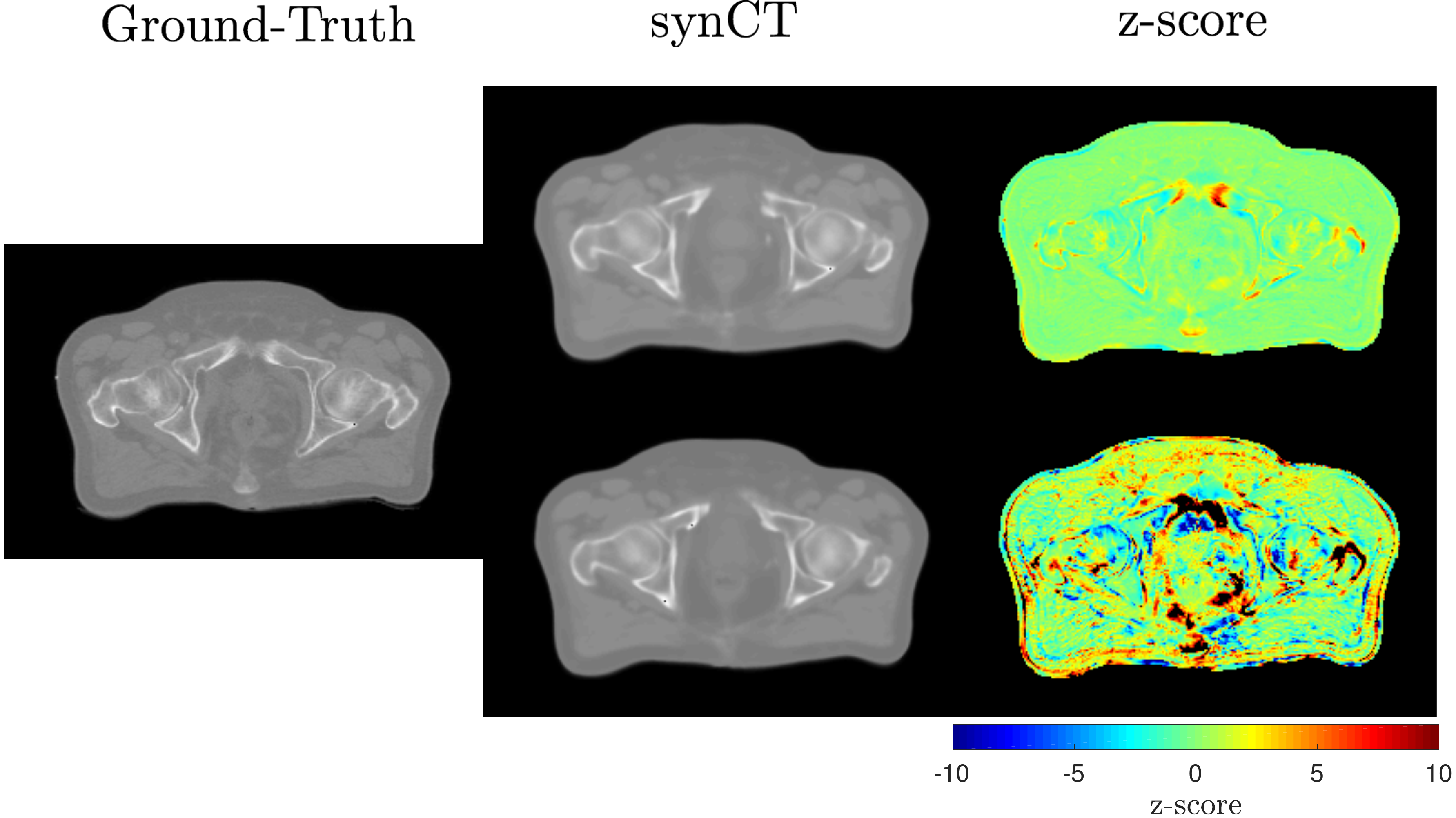}}
  \hspace{0.01\textwidth}
  {\includegraphics[trim=0mm 0mm 0mm 0mm,clip=true,height=0.29\textwidth,keepaspectratio]{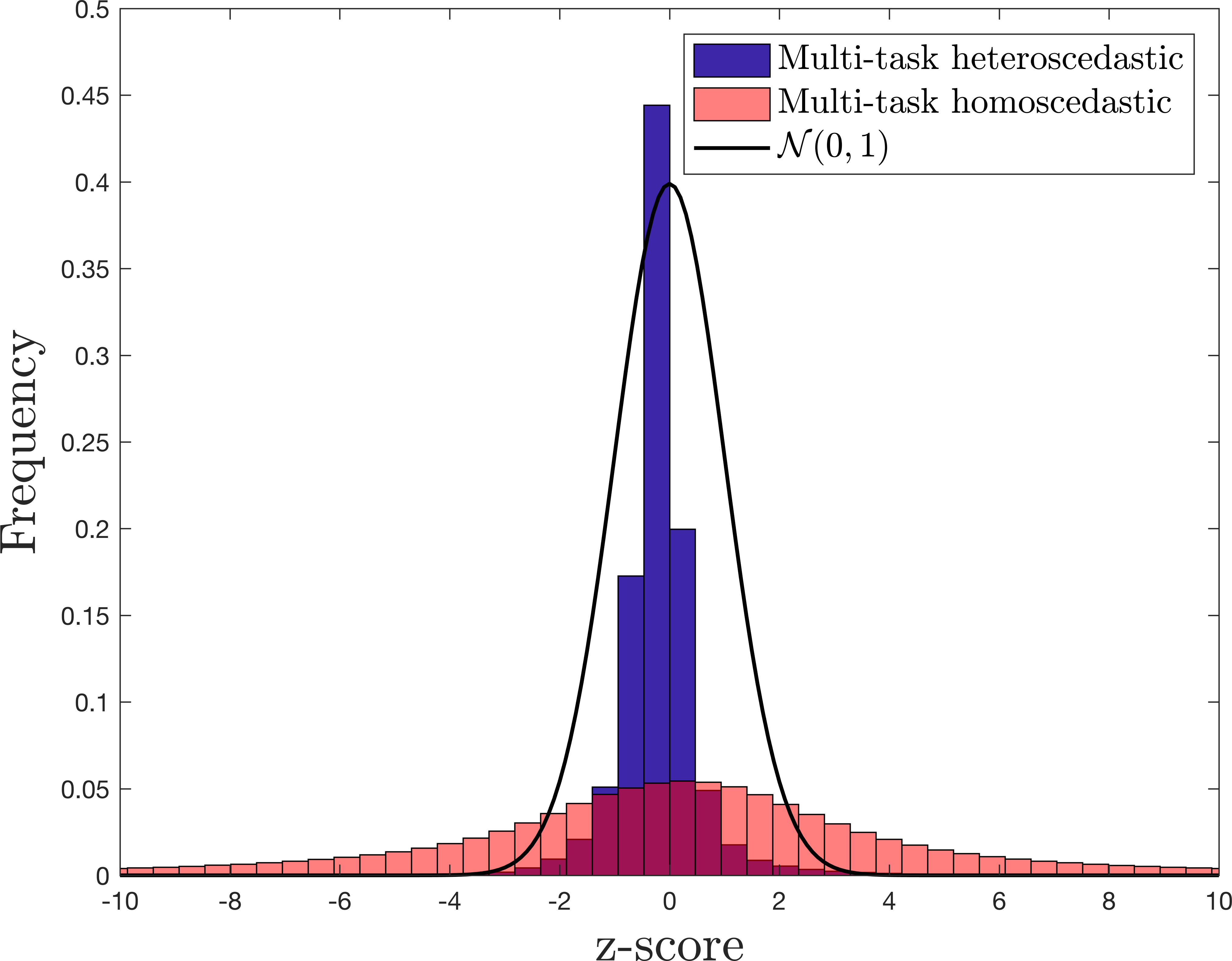}}
   \vspace{-1mm}
  \caption{Analysis of uncertainty estimation. a) synCTs and z-scores for the a subject between M4 (top) and M3 (bottom) models. b) z-score distribution of all patients ($15$) between both models.}
 \vspace{-2mm}
\end{figure}

\section{Conclusions}

We have proposed a probabilistic dual-network that combines uncertainty modelling with multi-task learning. Our network extends prior work in multi-task learning by integrating heteroscedastic uncertainty modelling to naturally weight task losses and maximize inductive transfer between tasks. We have demonstrated the applicability of our network in the context of MR-only radiotherapy treatment planning.  The model simultaneously provides the generation of synCTs, the segmentation of OARs and quantification of predictive uncertainty in both tasks. We have shown that a multi-task framework with heteroscedastic noise modelling leads to more accurate and consistent synCTs with a constraint on anatomical consistency with the segmentations. Importantly, we have demonstrated that the output of our network leads to consistent anatomically correct stochastic synCT samples that can potentially be effective in treatment planning. 

\subsubsection{Acknowledgements.} FB, JM, DH and MJC were supported by CRUK Accelerator Grant A21993. RT was supported by Microsoft Scholarship. ZER was supported by EPSRC Doctoral Prize. DA was supported by EU Horizon 2020 Research and Innovation Programme Grant 666992 and EPSRC Grant M020533, M006093 and J020990. We thank NVIDIA Corporation for hardware donation.

\begin{figure}[!t]
		\centering
		\includegraphics[height=0.14\textwidth]{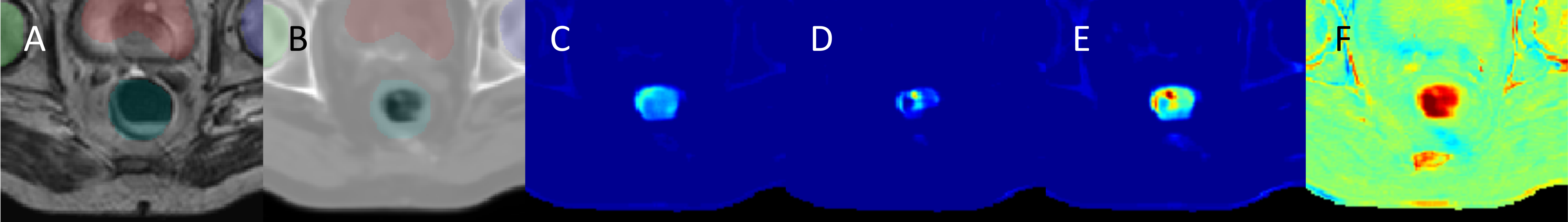}
		\caption{Uncertainty in problematic areas. a) T2 with reference segmentation, b) synCT with localised error, c) \textit{intrinsic} uncertainty, d) \textit{parameter} uncertainty, e) total predictive uncertainty and g) error in HU (range [-750HU, 750HU]).} 
		\label{fig:diagram4}
        \vspace{-5mm}
\end{figure}

\bibliographystyle{splncs}
\bibliography{bibliography}

\end{document}